\documentclass[sigconf]{acmart}
\AtBeginDocument{%
  \providecommand\BibTeX{{%
    \normalfont B\kern-0.5em{\scshape i\kern-0.25em b}\kern-0.8em\TeX}}}

\setcopyright{acmcopyright}
\copyrightyear{2021}
\acmYear{2021}
\setcopyright{acmlicensed}
\acmConference[SIGIR '21] {Proceedings of the 44th International ACM SIGIR Conference on Research and Development in Information Retrieval}{July 11--15, 2021}{Virtual Event, Canada.}
\acmBooktitle{Proceedings of the 44th International ACM SIGIR Conference on Research and Development in Information Retrieval (SIGIR '21), July 11--15, 2021, Virtual Event, Canada}
\acmPrice{15.00}
\acmISBN{978-1-4503-8037-9/21/07}
\acmDOI{10.1145/3404835.3463042}

\begin{document}
\fancyhead{}

\title{Towards an Online Empathetic Chatbot with Emotion Causes}

\author{Yanran Li*, Ke Li}
\authornote{Both authors contributed equally to this research.}
\author{Hongke Ning, xiaoqiang Xia, Yalong Guo, Chen Wei, Jianwei Cui, Bin Wang}

\affiliation{%
  \country{Xiaomi AI Lab}}
\email{liyanran,like16,ninghongke,xiaxiaoqiang,guoyalong,weichen,cuijianwei,wangbin11@xiaomi.com}

\renewcommand{\shortauthors}{Li, et al.}

\begin{abstract}
Existing emotion-aware conversational models usually focus on controlling the response contents to align with a specific emotion class, whereas empathy is the ability to understand and concern the feelings and experience of others. Hence, it is critical to learn the causes that evoke the users' emotion for empathetic responding, a.k.a. emotion causes. To gather emotion causes in online environments, we leverage counseling strategies and develop an empathetic chatbot to utilize the causal emotion information. On a real-world online dataset, we verify the effectiveness of the proposed approach by comparing our chatbot with several SOTA methods using automatic metrics, expert-based human judgements as well as user-based online evaluation.
\end{abstract}



\maketitle

\section{Introduction}
Since empathy plays a vital role for amicable social conversation and trustful social bonding~\cite{Decety2010TheNO}, it is critical to endow social chatbots with the ability of empathy. However, existing approaches tend to produce responses that are rarely empathetic, as indicated by previous studies~\cite{Rashkin2019TowardsEO}. Specifically, the majority of existing emotion-aware conversational models focus on controlling the response contents to align with a specific emotion class~\cite{Li2018ASC,Zhou2018EmotionalCM,Huang2018AutomaticDG,Shen2020CDLCD}, whereas empathy is the ability to understand and concern the feelings and experience of others~\cite{Eisenberg2009EmpathicRS}. 

To facilitate empathetic responding, we argue that it is also necessary for the chatbots to learn the causes behind user's emotion, a.k.a., emotion causes, in addition to the emotion class. Although emotion cause has been regarded as an important information for emotion classification in conversation~\cite{Gui2014EmotionCD,Poria2019EmotionRI,Poria2020RecognizingEC} and news text~\cite{Gui2016EmotionCE,Xia2019EmotionCausePE}, its benefit for empathetic response generation is yet to be explored. Especially for online chatbots, it is non-trivial to exploit the causal emotion information. Usually, open-domain chatbots are trained over ``pseudo'' conversations from social media platforms, 
where people are more reserved to share negative emotions as compared to private channels like messaging apps and chatbots~\cite{Goffman1981FormsOT,Bazarova2015SocialSO}. 
Though it is plausible to train empathetic chatbots using the crowd-sourced \textsc{EmpatheticDialogues} dataset~\cite{Rashkin2019TowardsEO}, the conversation behavior on it is still distinguished from what we observed from online real logs. To be specific, we sample 8,000 conversations in \textsc{EmpatheticDialogues} dataset and another 8,000 real online conversations from XiaoAI, one of the largest Chinese social chatbots,\footnote{According to Xiaomi’s second-quarter report, there are 78.4 million monthly active users of XiaoAI. \url{https://www.statista.com/statistics/967715/worldwide-xiaomi-ai-assistant-users/}} and manually label the emotion causes in the conversations. We find that while 89\% of conversations begin with the speaker expressing their emotion causes in \textsc{EmpatheticDialogues} dataset, there are only 7\% of online logs containing user emotion causes. In other words, online users tend not initiatively to self-disclose their emotion causes, and might not reveal the reasons if the chatbot did not explicitly ask. As a result, the chatbots trained over these datasets often simply produce general responses that are not empathetic, and exhibit detached to users, as shown in the upper example in Table~\ref{table:example}. This critically harms user experience and weakens the user-chatbot stickness. 

\begin{table}[!t]
\centering
\small
\caption{Two Conversation Examples.}
\begin{tabular}{l|l|c}
\hline
\textbf{Turn} & \textbf{Utterance} & \textbf{Strategy \& Cause}\\\hline
{U1} & {I'm upset.} & None \\
{S1} & {Everything will be OK.} & None \\\hline
{U1} & {I'm upset.} & None\\
{S1} & {Sorry to hear that. What happened?} & effective questioning \\
{U2} & {We \textbf{\emph{broke up}}}. & emotion cause \\
{S2} & Oh dear, it must be hurt. Did you & active listening \\
& argue for something? &  \\
\hline
\end{tabular}
\label{table:example}
\end{table}

To remedy these issues, we get inspired from Psychological studies~\cite{Karp2015CanEB,Lubis2019DialogueMA} and develop \textsc{Emma}, an online \textbf{Em}pathetic chatbot based on the user e\textbf{m}otion c\textbf{a}uses. 
When a user starts a conversation, our approach firstly detects user emotion class and recognizes the emotion causes. If no emotion cause is detected, our chatbot \textsc{Emma} directs users to self-disclose more based on effective questioning and active listening, the two counseling strategies that are often used by psychologists to encourage further engagement and gather detailed information during the counseling conversations~\cite{Comer2013ActiveEL,Bickmore2007PracticalAT}. Finally, \textsc{Emma} produces empathetic responses based on the conversation history, detected emotion class as well as the emotion causes. In brief, we highlight our contributions as follows: (1) We identify the significance of emotion cause for empathetic response generation; (2) We develop an online empathetic chatbot \textsc{Emma} by leveraging causal emotion information obtained through counseling strategies; (3) We curate a large-scale empathetic conversation dataset from real-world online logs, and manually annotate the emotion causes on them; (4) We empirically demonstrate the effectiveness of our approach using automatic metrics, expert-based human judgements as well as user-based online evaluation. The dataset and codes will be released soon. 

\section{Dataset: \textsc{X-EMAC}}
We construct a novel empathetic conversation dataset with causal emotion information to examine the benefit of emotion causes for empathetic response generation, especially in online environments. 

\noindent\textbf{Emotion Cause}. Following previous work~\cite{Poria2019EmotionRI,Poria2020RecognizingEC}, we define emotion cause as the continuous text span in an utterance that can be used to detected or inferred the speaker's emotion. 

To begin with, we randomly sample a large set of user queries from XiaoAi online logs, and ask human experts to annotate the queries with four common emotion classes: sad, anger, joy and others. These annotated queries are used to train an emotion classifier.

\noindent\textbf{Counseling Strategies}. To encourage online users to self-disclose more information, we hire psychologists to manually write a set of diverse templates using the counseling strategies of active listening and effective questioning~\cite{Comer2013ActiveEL,Bickmore2007PracticalAT}. The templates are designed specific to each emotion class and in average, there are 53 templates per class. Then, we deploy the templates online as the corresponding responses to those user queries that are classified into certain emotion classes (sad, anger, joy), and collect the next turn of real-time user responses to the templates. According to these three turns of conversations (online user-online template-online user), human experts are required to annotate the span of emotion causes in each utterance, and write empathetic responses with high-quality. We also filter dirty and sensitive conversations carefully. The statistics of the resulted \textbf{X}iaoAI \textbf{Em}pathetic \textbf{C}onversation (\textsc{X-EMAC}) dataset are summarized in below. 

\begin{table}[!h]
  \caption{Statistis of Dataset \textsc{X-EMAC}.}
  \label{tab:freq}
  \begin{tabular}{rc}
    \toprule
    Total Number of Conversations & 16,873 \\
    Total Number of Templates & 157\\
    Total Types of Emotion Causes & 29 \\
    Average Utterances per Conversation & 4\\
    Average Words per Utterance & 8.9\\
  \bottomrule
\end{tabular}
\end{table}

Based on the user experiences, the emotion causes in the annotated spans are manually categorized into 29 common and coarse-grained types like missing someone, broken up, etc. Notably, we find that 62\% of users will respond to our counseling-based templates, indicating the effectiveness of the proposed strategies in encouraging user engagement during online empathetic conversations, which will also be demonstrated in our experiments.

\section{Model: \textsc{Emma}}
\begin{figure}
  \includegraphics[width=\linewidth]{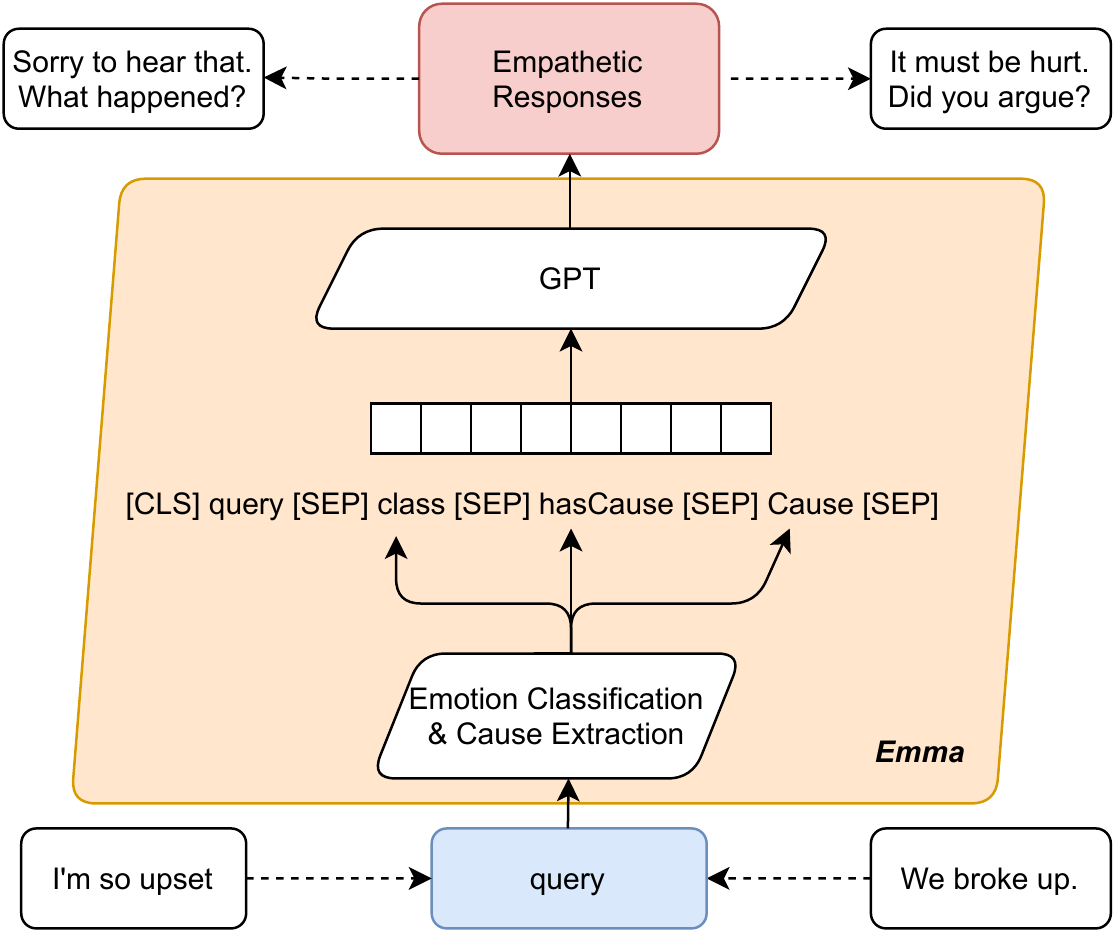}
  \caption{Emma: An Empathetic Chatbot.}
  \label{fig:emma}
\end{figure}

Formally, our task is to learn a response generation model via maximum likelihood estimation and generate a response $Y = \{y_1,\cdots,y_T\}$ according to the user query $X = \{x_1,\cdots,x_N\}$ and history conversations $H$, where $N$ and $T$ are the corresponding token numbers. To make the generated responses more empathetic, we also leverage emotional information like emotion label $L$ and emotion causes $C$. We will detail the methods of emotion classification and emotional cause extraction in the experiment parts. 
Overall, the conditional probability of generating a response $Y$ can be formulated as:
\begin{equation} 
\small
\begin{split}
P_Y = \prod_{i=1}^{T}P(y_t|y_{0:t-1}, X, H, L, C)
\end{split}
\label{eq:overall}
\end{equation}

Since pre-trained language models (PLMs) have demonstrated their great potentials in dialogue response generation~\cite{Zhang2020DialoGPTLG,Bao2020PLATOPD,Wolf2019TransferTransfoAT,Dong2019UnifiedLM,wang2020chinese}, we focus on the following sub-questions: (1) Whether PLMs equipped with causal emotion information are able to generate empathetic responses; (2) Whether PLMs are able to capture counseling strategies in leading empathetic conversations.

To answer these sub-questions, we develop an \textbf{Em}pathetic chatbot based on the user e\textbf{m}otion c\textbf{a}uses, namely \textsc{Emma}. Based on GPT~\cite{Radford2018ImprovingLU}, \textsc{Emma} also consists of multiple blocks of multi-head self-attention mechanism~\cite{Vaswani2017AttentionIA}. Specifically, we concatenate the embeddings of user query and emotional information into one, long text sequence. Following common practice~\cite{wang2020chinese}, we also include speaker embedding indicating speaker information and two special symbols [CLS] and [SEP] to separate the sequences.

As depicted in Figure~\ref{fig:emma}, when there is no history dialogue, the concatenated input to \textsc{Emma} contains the query embeddings, emotional label as well as emotion cause embeddings. Intuitively, \textsc{Emma} will produce responses with active listening and effective questioning strategies when the [hasCause] indicator is observed as [None]. As the conversation goes, \textsc{Emma} considers the history conversation information, and the input becomes [CLS][speaker1]q1[speaker2]r1 [speaker1]q2[sep]label[SEP]hasCause[SEP]Cause[SEP]. Due to space limit, we do not illustrate the history information in Figure~\ref{fig:emma}.

\section{Experiments}
To evaluate our approach \textsc{Emma} for building online empathetic chatbot, we conduct several sets of experiments\footnote{We will release the dataset and the codes for future research.} on the proposed dataset \textsc{X-EMAC} to examine: (1) the benefits of emotion cause; (2) the effectiveness of counseling strategies; and discuss (3) the difficulties and insights we learn.

\subsection{Experimental Setups}
\noindent\textbf{Compared Models}. We compare with the following approaches: (1) \textsc{BERT-Retrieve}, an 12-layer BERT~\cite{Devlin2019BERTPO} trained over a LCCC-large dataset,\footnote{\url{https://github.com/thu-coai/CDial-GPT}} which consists of 12 million multi-round short-text conversations from social media; (2) 
\textsc{UniLM-Generation}, the SOTA PLM for both natural language understanding and generation tasks with its official implementation;\footnote{\url{https://github.com/microsoft/unilm}} (3) \textsc{CDial-GPT}, the SOTA Chinese PLM for short-text conversation generation task with its official implementation;\footnote{\url{https://github.com/thu-coai/CDial-GPT}} (4) Our \textsc{Emma} and its variant \textsc{Emma}-Cause, where emotion cause information is removed from the input sequences.

\noindent\textbf{Implementation Details}. We divide the dataset into train/dev/test sets randomly with the ratio of 8:1:1, where dev set is used to tune the model parameters, and the test set is used to compare the model performances. For word segmentation, we use Jieba,\footnote{\url{https://github.com/fxsjy/jieba}} and the vocabulary is constrained to 1,998 words. The dimension of word embedding is set to 768, the number of the warmup epoch is set to 1, and the maximum learning rate is 5e-3. We implement the models based on the huggingface library,\footnote{\url{https://github.com/huggingface/transformers}} and test them on NVIDIA Telsa V100. For fair comparison, all the compared models are using 12 layers with 12 attention heads, and optimized using AdamW optimizer~\cite{Loshchilov2017FixingWD}. Based on the empirical results, the number of gradient accumulation is set to 64.

\noindent\textbf{Evaluation Metrics}. We adopt three kinds of assessments: (1) The \textbf{word-level automatic metrics} including Perplexity (PPL) and distinct n-grams~\cite{Li2016ADO}, which indicate the fluency and informativeness of the generated responses respectively; (2) The 3-scale \textbf{expert-based human judgements} in terms of empathy and relevance~\cite{Rashkin2019TowardsEO}. We randomly sample 250 conversations and the responses from each compared model. Then, we hire 3 professional experts from a third-party annotation company to evaluate the response quality with scale of \{0,1,2\}. For example, empathy measures the extent of understanding exhibited by the chatbots to the users' feelings; (3) A \textbf{user-based online evaluation} net sale value (NSV), which suggests the users preferences on chatbot's replies. In online chatbots and social platforms like Reddit, users can click the icons of thumb up and down, indicating they like or dislike a response. By summing up the total numbers of ``like'' and ''dislike'' a chatbot (method) receives, we are able to calculate that $\text{NSV} = (\#upvotes-\#downvotes) \/ (\#upvotes+\#downvotes)$. The higher the NSV is, the better the conversational model is preferred.

\subsection{Results and Analysis}
\noindent\textbf{The Benefit of Emotion Causes}. As shown in Table~\ref{tab:exp_main}, the retrieval-based BERT model in the first row performs the worst. Even though it yields decent distinct-n scores and satisfactory relevance score, its responses are rated as the least empathetic by human experts and are disliked most by the online users. Considering that \textsc{BERT-Retrieve} is trained on 12 million of conversations collected from internet, its disappointing performance validates that it is unsafe to directly train online open-domain chatbot using social media data, which may contain a lot of less empathetic responses that will harm user experience.

The performances of \textsc{UniLM-Gen} and \textsc{CDial-GPT} in the second and third rows are similar but defective. They are SOTA response generation models and are expected to capture implicit knowledge like conversation patterns~\cite{Li2020AnEI}. However, as implied by the expert-based judgements, the responses generated by them are of low empathy and moderate relevance. Between them, \textsc{UniLM-Gen} is slightly better than \textsc{CDial-GPT}. We conjecture that it is because \textsc{UniLM-Gen} together models the bi-directional context, while \textsc{CDial-GPT} only follows a left-to-right manner to predict a text sequence~\cite{Dong2019UnifiedLM}. Nevertheless, the online user feedback on \textsc{UniLM-Gen} still suggests a room for improvement.

\begin{table}
  \caption{Experimental Results on Using Emotion Cause.}
  \label{tab:exp_main}
  \begin{tabular}{lccccc}
    \toprule
    Model & PPL & Dist-1  & Empathy & Relev. & NSV \\
    \midrule
    \textsc{BERT-Ret}&- &\textbf{0.291}&0.97&1.69& -3.1\%\\
    \textsc{UniLM-Gen} &58.54&0.138 & 1.03&1.68& 4.3\% \\ 
    \textsc{CDial-GPT} &62.33&0.083&1.03&1.42& - \\
    \textsc{Emma} &\textbf{19.66} &0.039&\textbf{1.43}&\textbf{1.75}& \textbf{28.5\%}\\\midrule
    \multicolumn{1}{r}{-Cause} &20.79&0.031&1.24&1.59&16.9\% \\
  \bottomrule
\end{tabular}
\end{table}

\begin{table}
  \caption{Human A/B Test Results.}
  \label{tab:exp_ab}
  \begin{tabular}{lccc}
    \toprule
    Model & Win & Loss & Tie\\
    \textsc{Emma} v.s. \textsc{BERT-Ret} & 56\% & 6\% & 37\%\\
    \textsc{Emma} v.s. \textsc{UniLM-Gen} &51\%& 15\% & 34\%\\
    \textsc{Emma} v.s. \textsc{CDial-GPT}  & 62\% & 6\% & 32\%\\\hline
    \textsc{Emma} v.s. \textsc{Emma-Cause} & 35\%  & 12\% & 53\%\\
  \bottomrule
\end{tabular}
\end{table}

When examining the performances of \textsc{Emma} and its variant (the last two rows), we can see that the proposed approach outperforms the other PLMs especially on empathy and NSV scores. The improvements are largely brought by emotional information utilization, which demonstrates the significance of emotional information in empathetic response generation. The performance gap between \textsc{Emma} and \textsc{Emma-Cause} further indicates the benefit of emotion cause, which has been neglected before. While emotion class label is integrated into both \textsc{Emma} and its variant \textsc{Emma-Cause}, only \textsc{Emma} has the ability of capturing fine-grained user experience and responding with proper communication strategies, the ability of empathy~\cite{Eisenberg2009EmpathicRS}. 

We also carry out an expert-based human A/B test to verify the comparison results. We compare the best model \textsc{Emma} with each other PLM in a pair-wise manner. For fair comparison, the expert annotators do not know which model the response is from. As shown in Table~\ref{tab:exp_ab}, it is obvious that our empathetic chatbot \textsc{Emma} are more favored by the professionals, which is consistent with online user preferences.

\noindent\textbf{The Effectiveness of Counseling Strategies}. The experimental results shown above also imply that PLM models like GPT are able to learn counseling strategies on the \textsc{X-EMAC} dataset. Typically, we leverage the strategies of effective questioning and active listening, which enable the chatbot to sound more attentive and encourage the users to describe more about their situations~\cite{Lubis2019DialogueMA}. It is thus possible to apply these strategies for acquiring more detailed information in the second-round of user queries than their initial ones.

\begin{table}
  \caption{Experimental Results on Counseling Strategies (CS).}
  \label{tab:exp_cs}
  \begin{tabular}{lcccc}
    \toprule
    Model & Dist-1 & Dist-2& Empathy & Relevance \\
    \midrule
    \textsc{BERT-Retrieve}& \textbf{0.294}&\textbf{0.596} &0.95&1.60\\
    \textsc{BERT-Retrieve}+CT. & 0.106&0.254&1.06&\textsc{1.77}\\
    \textsc{UniLM-Gen} & 0.140&0.402&1.02&1.58\\
    \textsc{UniLM-Gen}+CT. &0.036&0.103&\textbf{1.17}&1.69\\
  \bottomrule
\end{tabular}
\end{table}

\begin{table}
  \caption{Case Studies. Due to space limit, we omit the original Chinese text.}
  \label{tab:exp_case}
  \begin{tabular}{ll}
    \toprule
    Context/Model & Query/Responses \\\midrule
    \multicolumn{1}{r}{\emph{q1:}} & \emph{I am unhappy.} \\
    \multicolumn{1}{r}{\emph{r1-CS:}}& \emph{Don't worry, I'm always here for listening.}\\
    \multicolumn{1}{r}{\emph{q2:}}& \emph{My father beat me.} \\
    \midrule
    \textsc{BERT-Ret.} & I'm also unhappy this year. \\
    \textsc{BERT-Ret.}+CT. & What happened? \\
    \textsc{UniLM-Gen} & Just leave it.\\
    \textsc{UniLM-Gen}+CT. & He beat you because you did not listen to him.\\
    \textsc{CDial-GPT}& My father and mother also beat me.\\
    \textsc{Emma}& Sounds terrible! Could you tell more about why \\
    & and when it happened? \\
  \bottomrule
\end{tabular}
\end{table}

To this end, we conduct another set of experiments to investigate the strategy effectiveness. We take \textsc{BERT-Retrieve} and \textsc{UniLM-Generation} as exemplars, and compare their responses to two different set of queries. The first set of queries are the users' beginning queries as they start the conversations, and the other set are the second-round queries as the users replying to the templates with counseling strategies (CS). The comparison results are summarized in Table~\ref{tab:exp_cs}, where we can see clear improvements over empathy and relevance when both model are fed with more specific queries acquired through counseling strategies. 

When examining the distinct-n scores in Table~\ref{eq:overall} and Table~\ref{tab:exp_cs}, it is also worth noting that the distinct-n scores seem often contradict to expert-based and user-based judgements, which has also been found in previous studies~\cite{Sharma2021TowardsFE,welivita-pu-2020-taxonomy}. After manually checking the examples of generated responses from the compared models, we conclude that the responses with too many distinct n-grams often include content that are not sympathetic to users experiences. 
This somehow results in a decrease of response diversity. For better understanding, we show some examples of generated responses from the compared models in Table~\ref{tab:exp_case}.

\subsection{Discussions}

\noindent\textbf{The Mutual Benefits of Joint Learning.} There exist auxiliary approaches for the tasks of emotion classification (ECf.) and emotion cause extraction (ECE)~\cite{Poria2019EmotionRI}. In this work, we cast ECf. as a multi-classification task, and ECE as a reading comprehension task~\cite{Rajpurkar2016SQuAD10}. Since these two tasks are standard and dominated with PLM models, we adopt BERT to perform both ECf. and ECE. 

\begin{table}
  \caption{Results on Emotion Classification (ECf.) and Emotion Cause Extraction (ECE).}
  \label{tab:exp_ece}
  \begin{tabular}{rcccc}
    \toprule
    Task & Precision & Recall & Exact\_Match & Fuzzy\_Match \\\midrule
    ECf. &0.924 & 0.862 & - & - \\
    ECE & - & - & 0.746 & 0.903\\ 
    \textbf{ECf. + ECE} & \textbf{0.930} & \textbf{0.863} & \textbf{0.750} & \textbf{0.910} \\
  \bottomrule
\end{tabular}
\end{table}

Following common practice, we report the precision and recall for ECf., whereas Exact\_Match and Fuzzy\_Match scores for ECE in Table~\ref{tab:exp_ece}. Because we only have 4 emotion classes annotated in \textsc{X-EMAC} dataset, it is fair that BERT-based model yields acceptable precision and recall scores. Clearly, the scores on emotion cause extraction task (ECE) are lower than those on emotion classification (ECf.). It is reasonable because ECE is a more difficult task than ECf. As discussed in~\cite{Poria2020RecognizingEC}, emotion causes can be context-depedent or context-independent, and sometimes they are latent and unmentioned in the text. Inspired by previous work~\cite{Chen2018JointLF,Xia2019EmotionCausePE}, we add up the losses from the two tasks and joint learning the model parameters. Despite the fact that both tasks are inherited with some difficulties, these two tasks are mutually indicative for each other. By combining the two worlds, we are able to further improve the understanding of the conversation and increase the precision and accuracy of the tasks.

\noindent\textbf{Error Analysis}. It is also promising to jointly learn the emotional understanding model and response generation model for better holistic performance. Currently, we do not regulate the generated responses and the model is still possible to bypass the causal emotion information during the generation. Intuitively, we can perceive the empathy level of the generated responses and formulate it as an additional loss to be optimized together. Such loss is beneficial to check whether the user emotion is well understood, and is also helpful for generation model to improve itself. In this way, we are able to build up the empathetic chatbots in a holistic fashion.

\noindent\textbf{Future Work}. The focus of this work is to explore the benefit of emotion cause in empathetic response generation using a simple but effective approach. Towards building building online empathetic chatbots, there are other important research problems such as how to mimicking user's emotion more naturally~\cite{Majumder2020MIMEME} and how to integrating commonsense knowledge more selectively~\cite{Li2020TowardsED}. In the future, we plan to improve the understanding of user emotion by integrating commonsense reasoning, and investigate response rewriting to make the counseling templates more natural.

\section{Ethics Declaration}
The dataset was sourced with license and consent from XiaoAI chatbot, and this work was approved by XiaoAI team. All personally identifiable information in our dataset was removed. Towards preventing unsafe expressions, all of our responses and strategies are proof-read and double-checked by XiaoAI team and mental health professionals before online deployment.

\bibliographystyle{ACM-Reference-Format}
\bibliography{sample-base}
\end{document}